\documentclass{article}
\usepackage{ijcai26}

\usepackage{times}
\usepackage{soul}
\usepackage{url}
\usepackage[hidelinks]{hyperref}
\usepackage[utf8]{inputenc}
\usepackage[small]{caption}
\usepackage{graphicx}
\usepackage{amsmath}
\usepackage{amsthm}
\usepackage{booktabs}
\usepackage{algorithm}
\usepackage[noend]{algorithmic}
\usepackage[switch]{lineno}
\usepackage{adjustbox}

\usepackage{amsfonts}
\usepackage[caption=false,font=normalsize,labelfont=sf,textfont=sf]{subfig}
\usepackage{textcomp}
\usepackage{stfloats}
\usepackage[table]{xcolor} 
\usepackage{amssymb}
\usepackage{pifont}
\usepackage{multirow}

\title{CogMCTS: A Novel Cognitive-Guided Monte Carlo Tree Search Framework for Iterative Heuristic Evolution with Large Language Models}

\author{
    Author Name
    \affiliations
    Affiliation
    \emails
    email@example.com
}

\author{
Hui Wang$^1$$^2$
\and
Yang Liu$^1$$^2$\and
Xiaoyu Zhang$^1$$^2$\And
Chaoxu Mu$^1$$^2$$^3$\\
\affiliations
$^1$School of Artificial Intelligence, Anhui University\\
$^2$Anhui Provincial Key Laboratory of Security Artificial Intelligence, Anhui University\\
$^3$Pengcheng Laboratory, Shenzhen, China\\
\emails
\{h.wang.13, zhangxiaoyu\}@ahu.edu.cn,
wa24201028@stu.ahu.edu.cn,
cxmu@tju.edu.cn
}
\begin{document}

\maketitle

\begin{abstract}
Automatic Heuristic Design (AHD) is an effective framework for solving complex optimization problems. The development of large language models (LLMs) enables the automated generation of heuristics. Existing LLM-based evolutionary methods rely on population strategies and are prone to local optima. Integrating LLMs with Monte Carlo Tree Search (MCTS) improves the trade-off between exploration and exploitation, but multi-round cognitive integration remains limited and search diversity is constrained. To overcome these limitations, this paper proposes a novel cognitive-guided MCTS framework~(CogMCTS). CogMCTS tightly integrates the cognitive guidance mechanism of LLMs with MCTS to achieve efficient automated heuristic optimization. The framework employs multi-round cognitive feedback to incorporate historical experience, node information, and negative outcomes, dynamically improving heuristic generation. Dual-track node expansion combined with elite heuristic management balances the exploration of diverse heuristics and the exploitation of high-quality experience. In addition, strategic mutation modifies the heuristic forms and parameters to further enhance the diversity of the solution and the overall optimization performance. The experimental results indicate that CogMCTS outperforms existing LLM-based AHD methods in stability, efficiency, and solution quality.
\end{abstract}

\section{Introduction}

Heuristic algorithms have shown strong potential to solve complex search problems and NP-hard combinatorial optimization problems (COPs) ~\cite{desale2015heuristic}. They are widely applied in real-world scenarios, including logistics ~\cite{tresca2022automating}, scheduling ~\cite{palacio2022q}, traffic control ~\cite{he2011heuristic}, and robotics ~\cite{tan2021comprehensive}. Over the past decades, researchers have devoted considerable effort to designing heuristic methods. Classical metaheuristic algorithms, such as simulated annealing ~\cite{kirkpatrick1983optimization}, tabu search ~\cite{glover1990tabu,glover1990tabu1}, and iterative local search ~\cite{lourencco2003iterated}, have been proposed. However, different applications often involve heterogeneous constraints and objectives. This usually requires tailored algorithms or specific configuration strategies. Manual design, tuning, and optimization of heuristics is complex and time-consuming. It also depends heavily on domain knowledge and experience. Consequently, these approaches face obvious limitations in scalability and generality.

To achieve a more efficient and convenient heuristic design in all tasks, researchers have proposed the concept of Automatic Heuristic Design~(AHD)~\cite{burke2013hyper}. Its main goal is to automatically generate, adjust, or optimize heuristics for specific problem types, improving the efficiency and adaptability of the solution. In this framework, representative approaches include Genetic Programming (GP) ~\cite{langdon2013foundations}, Neural Combinatorial Optimization (NCO) ~\cite{qu2020general}, and Hyper-Heuristics (HHs) ~\cite{pillay2018hyper}. However, these methods have certain limitations. GP-based AHD relies on a manually predefined set of operators. NCO requires appropriate inductive bias and faces challenges in interpretability and generalization ~\cite{liu2023good}. HHs remain constrained by predefined heuristics by human experts.

In recent years, large language models(LLMs) have shown strong abilities in many domains ~\cite{naveed2025comprehensive}, opening new directions for AHD. Studies show that LLMs ~\cite{nejjar2025llms} can be powerful tools to generate new ideas and heuristic algorithms. LLM-based AHD methods ~\cite{liu2024systematic} can build high-quality heuristics for complex problems without human help. They improve efficiency and adaptability in algorithm design. Recently, several studies have combined LLMs with Evolutionary Computation (EC) to support automatic heuristic generation. These approaches are often called LLM-based Evolutionary Program Search (LLM-EPS) ~\cite{liu2024large,meyerson2024language,chen2023evoprompting}. Early work such as FunSearch ~\cite{romera2024mathematical} and later methods including Evolution of Heuristic (EoH) ~\cite{liu2024evolution}, Reflective Evolution (ReEvo) ~\cite{ye2024reevo}, and HSEvo ~\cite{dat2025hsevo} demonstrated strong performance in producing effective heuristics. These methods rely on population-based optimization. Low-performing heuristics are eliminated and the search is directed toward high-quality candidates. This process improves the overall efficiency. During this process, some methods introduce reflection mechanisms. These mechanisms iteratively evaluate generated heuristics and accumulate experience to guide subsequent search. However, weak heuristics may still evolve into promising solutions under iterative optimization with LLMs. 
Therefore, a pure population structure risks premature convergence and restricts exploration of heuristic space. To overcome this issue, methods such as MCTS-AHD ~\cite{zheng2025monte} and Planning of Heuristics (PoH) ~\cite{wang2025planning} integrate LLMs with Monte Carlo Tree Search(MCTS). The use of the UCT strategy provides a balance between exploration and exploitation, reducing the risk of local optima in population-based methods. Despite this progress, several limitations remain. The reflection mechanisms introduced are often limited to single-round or shallow evaluations. Such mechanisms make it difficult to combine past experience with current heuristics across multiple iterations. Node expansion strategies are mostly single-path, which limits diversity and exploration. In addition, the reflection and search processes are weakly connected. As a result, some promising heuristics may be overlooked, reducing overall optimization effectiveness.

To address these limitations, we propose a novel framework that integrates a cognitive guidance mechanism into the MCTS search. The framework achieves efficient and systematic heuristic optimization. We keep an eye on high-performing heuristics. It also allows low-performing ones to improve through multi-round cognitive guidance from the LLM. This enables a broader exploration of the heuristic space. The integration of fast cognition and complex cognition in multiple rounds gives structured feedback to the LLM. Feedback includes historical experience, current node information, and negative outcomes. This feedback improves both the relevance and the effectiveness of heuristic generation. Dual-track node expansion with UCT selection and an elite heuristic set maintains a balance between exploration and the use of high-quality experience. This balance reduces the risk of premature convergence. Cognitive-guided strategic generation introduces new changes in heuristic forms and parameter settings. These changes improve both the diversity and the quality of solutions. The experimental results show that the framework is more robust and efficient than the existing LLM-based AHD methods. It sustains high-reward heuristic optimization and uncovers potential high-quality solutions. The results show significant advantages in solving complex combinatorial optimization problems.

Overall, the main contributions of this paper can be summarized as follows:
\begin{enumerate}
    \item We propose a multi-round cognitive guidance mechanism. It enhances high-performing heuristics through fast and complex cognition and refines low-performing ones. This design improves the relevance and effectiveness of heuristic generation.
    \item We design a dual-track node expansion and structured feedback strategy. It maintains a balance between exploration and exploitation, improving the quality and robustness of the generated heuristics.
    \item The proposed method is implemented within the Ant Colony Optimization (ACO), Guided Local Search (GLS), and Step-by-Step frameworks. It successfully solves diverse combinatorial optimization problems and achieves state-of-the-art performance across multiple tasks, demonstrating strong adaptability and superiority.
\end{enumerate}

\section{Related Work}
\subsection{Automatic Heuristic Design}

AHD is a general approach for COPs. Its goal is to systematically generate or select efficient heuristic algorithms. AHD does not follow traditional manual design. It searches for the best strategies in a fixed heuristic space ~\cite{drake2020recent}. It can also create new strategies by joining basic heuristic components ~\cite{duflo2019gp,zhao2023automated}. This process reduces the need for human intervention. In practice, each heuristic can be treated as an algorithm that maps problem instances to feasible solutions. For example, in the 0-1 Knapsack Problem (KP), the input includes item weights, values, and knapsack capacity. The heuristic outputs a selected set of items that maximizes the total value without exceeding the capacity. To evaluate heuristic performance, AHD methods typically assess the solutions generated in task-specific datasets and compute performance metrics based on the quality of the solution. Heuristics can be incorporated into different general frameworks, and the optimal strategy may exist in multiple frameworks. In stepwise solution construction frameworks, KP heuristics build the final solution by selecting items one by one. The quality of the heuristic function directly affects overall solution quality and construction efficiency. However, traditional AHD methods still rely on extensive domain knowledge and complex implementations. This dependency limits the full automation potential of AHD ~\cite{o2010open,pillay2018hyper}.
\subsection{LLM-driven AHD}
The development of LLMs has brought new opportunities to multiple fields ~\cite{xi2025rise,wang2024survey,zhang2024proagent,lu2023medkpl}. LLMs have been widely used for code debugging ~\cite{chen2023teaching,liventsev2023fully}, algorithm contest solving ~\cite{shinn2023reflexion}, robotics ~\cite{lehman2023evolution,liang2022code,wang2023gensim}, performance optimization ~\cite{shypula2023learning}, and general task solving ~\cite{yang2023intercode}. LLMs are increasingly regarded as powerful tools for designing novel heuristic algorithms. Relying only on prompt engineering ~\cite{yang2023large,wei2022chain} cannot surpass the limits of existing knowledge and cannot continuously produce novel and effective heuristics. To address this limitation, researchers have introduced LLMs into evolutionary algorithms ~\cite{yang2023large,wu2024evolutionary,li2024bridging,hemberg2024evolving}, forming a hybrid optimization paradigm that integrates language models and evolution approaches. The FunSearch framework from Google DeepMind ~\cite{romera2024mathematical} generates code snippets iteratively with LLMs and evaluates them using task-specific reward functions. Successful candidates are fed back to guide further generation. Later, Liu et al. ~\cite{liu2024evolution} proposed the EoH framework, combining LLM generation with evolutionary mechanisms to propose new heuristics and iteratively improve them, creating a closed-loop optimization system. However, traditional evolutionary computation still faces challenges in maintaining diversity and utilizing historical information. To address this, Zhi et al. ~\cite{zheng2025monte} proposed the MCTS-AHD framework, which integrates MCTS with LLMs. The framework applies the UCT strategy of MCTS to balance exploration and exploitation. Low-performing heuristics can still evolve, which increases search diversity and prevents premature convergence.

\begin{figure*}[t]
  \centering
  \adjustbox{trim=1.5cm 1.5cm 0.5cm 1.3cm,clip}{
    \includegraphics[width=1.0\textwidth]{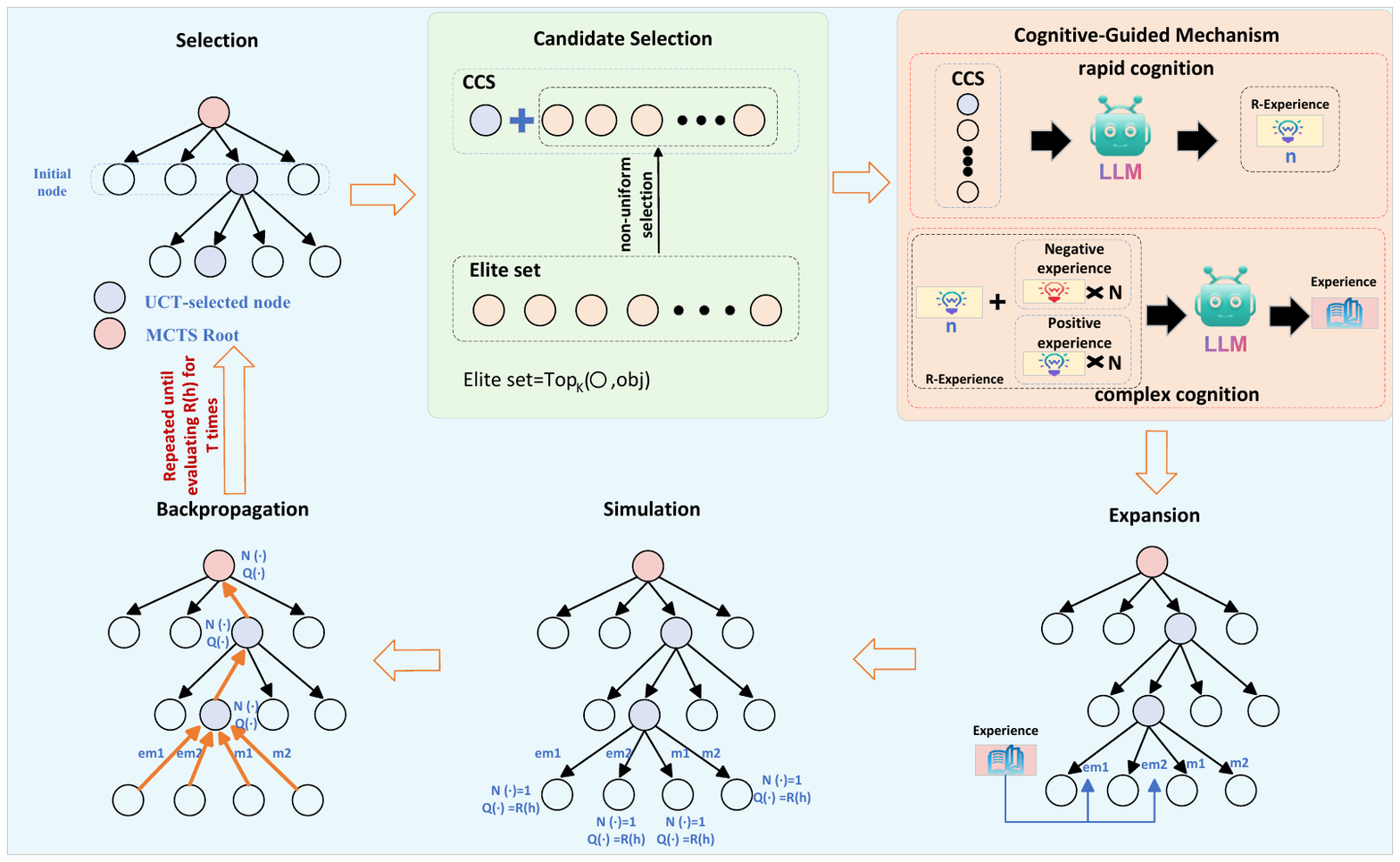}
  }
  \caption{CogMCTS framework diagram. It shows the close integration of the cognitive guidance mechanism with the MCTS process. This integration enables intelligent guidance of heuristic generation during the search.}
  \label{fig:vis1}
\end{figure*}

\subsection{LLM Self-Reflection in Heuristic Generation}

Despite significant progress in heuristic generation using hybrid optimization frameworks, these methods still rely mainly on external evaluation mechanisms. To improve generation quality, researchers have begun to explore the self-reflection ability of LLMs. Self-reflection allows the model to actively assess and optimize its outputs during generation. It is a cognitive process. The model reviews its reasoning, actions, and experience. It identifies inefficient or incorrect strategies. The model adjusts these strategies accordingly. Guiding LLMs to perform self-reflection can effectively enhance heuristic generation and problem-solving performance ~\cite{shinn2023reflexion,kumar2024supporting,huang2022inner}. Ye et al. ~\cite{ye2024reevo} proposed the ReEvo framework. It combines short-term and long-term self-reflection with evolutionary search. Short-term reflection generates offspring heuristics. Long-term reflection collects historical experience. It forms improvement suggestions and allows for knowledge accumulation and optimization. Heuristics are represented as code snippets and undergo systematic iterative optimization through population initialization, selection, reflection, crossover, and elite mutation. Similarly, Dat et al. ~\cite{dat2025hsevo} proposed the HSEvo framework. Rapid reflection is used to analyze parent combinations. Historical information from previous steps is used to produce immediate improvement guidance. Evolutionary strategies can be adjusted dynamically. Hence, search efficiency improves and individual improvements become more targeted. Building on these ideas, Wang et al. proposed the PoH ~\cite{wang2025planning} framework, combining LLM self-reflection with MCTS. The PoH framework systematically explores the heuristic space. It performs self-evaluation and improvement after generation. Therefore, iterative optimization is achieved and high-quality heuristics are discovered more efficiently. In contrast, our framework further strengthens the integration of cognitive guidance and search. Dual-track node expansion (\textit{em1}/\textit{em2} and \textit{m1}/\textit{m2}) is used. Multi-round fast and complex cognition is combined. Historical experience, current node information, and negative feedback are systematically passed to the LLM. This significantly improves the relevance and robustness of heuristic generation. In addition, the UCT strategy guides the selection of nodes, maintaining a balance between exploration and the use of high-quality historical experience. This enables more robust and efficient iterative improvement, enhances search diversity, and improves solution quality, outperforming existing frameworks.

\section{Preliminary}
\subsection{AHD for Combinatorial Optimization}

A COP is defined by a solution space $\mathcal{S}$ and an objective function $f: \mathcal{S} \to \mathbb{R}$. 
The goal is to find a solution in the space that optimizes the objective function. 
In the study of heuristics, we introduce a heuristic space $\mathcal{H}$. 
Each heuristic $h \in \mathcal{H}$ defines a mapping from an input instance to a candidate solution. 
The aim of heuristic optimization is to find the best heuristic $h^{*}$ that minimizes the meta-objective function $F: \mathcal{H} \to \mathbb{R}$, that is:
\begin{equation}
h^{*} = \arg\min_{h \in \mathcal{H}} F(h).
\end{equation}
Unlike traditional methods that rely on manual design or limited search, this study uses LLMs to generate heuristics directly. This allows for the exploration of a potentially infinite and open heuristic space. The approach goes beyond fixed rules and predefined operators, giving more flexibility and diversity to heuristic generation. In practice, the generated heuristics are evaluated on a task-specific set of training instances $\mathrm{I}$. To measure effectiveness, we define a reward function $\mathrm{R}$. The reward is computed from the performance of the heuristic in the set of instances. The final goal is to find the heuristic with the best performance under this evaluation mechanism. The optimization process can be written as:
\begin{equation}
h^{*} = \arg\max_{h \in \mathcal{H}} R(p_B(I,h)).
\end{equation}
\subsection{Self-reflection}
In LLM-based AHD, Self-reflection is the process through which a model iteratively improves its heuristics. This process relies on internal reasoning and evaluation. Let $H=\{h_1,h_2,\ldots,h_N\}$ denote the set of heuristics and $I$ denote the set of problem instances. The effectiveness of each heuristic $h\in H$ can be quantified by a reward function $R(h,I)$. The self-reflection process can be formalized as a mapping:
\begin{equation}
f: H \times I \rightarrow S
\end{equation}
Where, $S$ represents the structured feedback space. The mapping $f$ is driven by the LLM. It generates feedback $f(h, I) \in S$ by analyzing the logic, structure, and suitability of $h$ for $I$ . The Feedback may include natural language explanations, pseudocode modification suggestions, or performance scores. The feedback $f(h,I)$ guides the LLM to modify and optimize the original heuristic $h$. This process iteratively produces a higher-performance heuristic set $H'$.
\subsection{Evolution}
Evolution refers to the process of generating new heuristics by modifying, recombining, or varying parameters of existing heuristics. Let $P_t = \{h_1, h_2, \dots, h_N\}$ denote the population of heuristics at iteration $t$. The evolutionary process applies operators such as selection, recombination, and mutation to produce the next generation $P_{t+1}$:
\begin{equation}
P_{t+1} = E(P_t)
\end{equation}
Where $E$ represents the evolutionary operator. Each heuristic $h \in P_t$ in the population is evaluated using the reward function $R(h,I)$, which measures the effectiveness of the heuristic $h$ on the instance set $I$.
\begin{figure*}[t]
  \adjustbox{trim=1.0cm 3.6cm 0.5cm 1.0cm, clip}{
\includegraphics[width=1.0\textwidth]{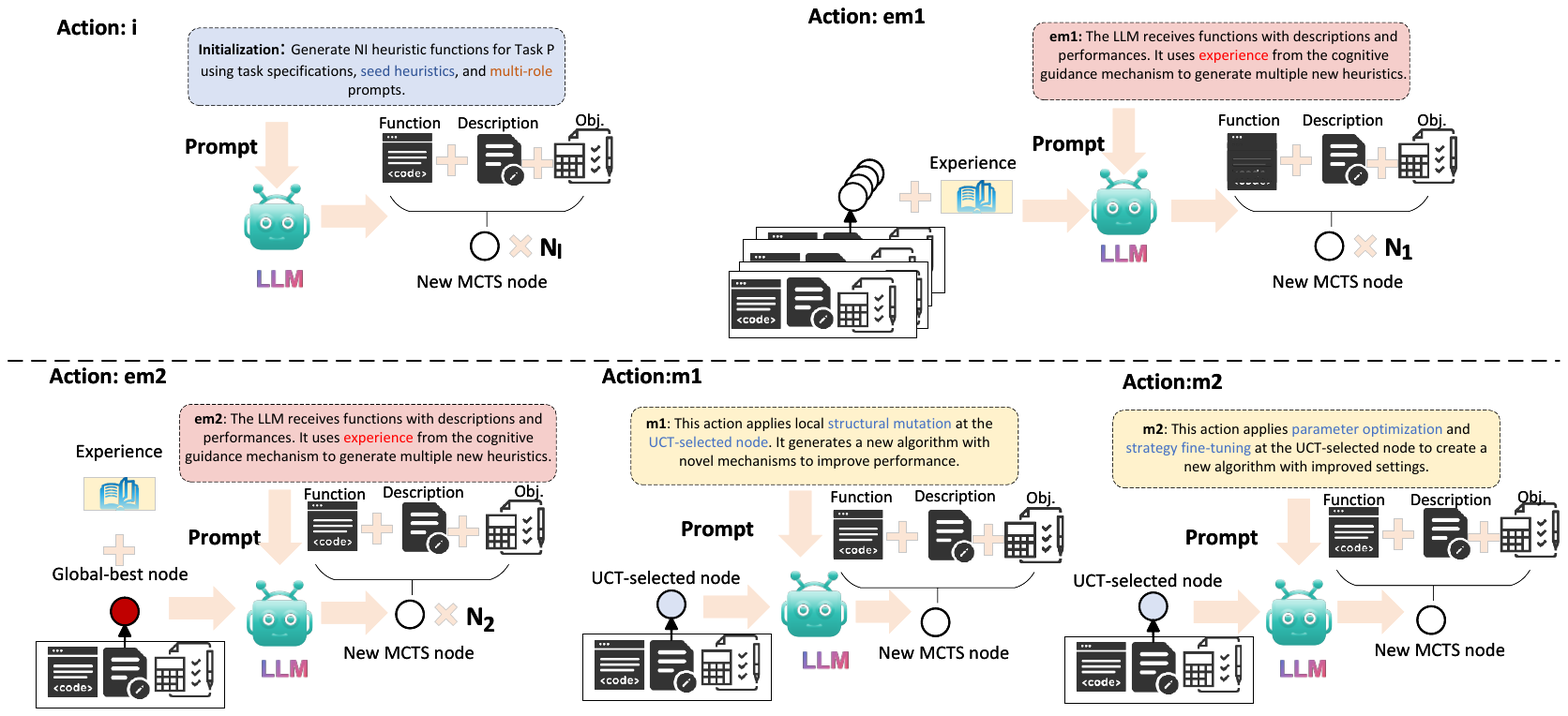}
    }
  \caption{In CogMCTS, LLM-based heuristic evolution includes three types of actions. First, initialization creates new heuristics (i). Second, cognitive-guided combination generates improved heuristics from multiple functions (em1, em2). Third, local mutation modifies existing heuristics to add diverse mechanisms or refined designs (m1, m2). }
    \label{fig:vis2}
\end{figure*}
\begin{algorithm}[t]
    \caption{CogMCTS}
    \label{algorithm}
    \textbf{Input}: Dataset $D$\\
    \textbf{Parameter}: Init nodes $N_I$, max evals $T$, elite set $E$, depth $H{=}10$, actions $\{i,em1,em2,m1,m2\}$, UCT $\lambda_0{=}0.1$, set size $\text{real\_m}{=}7$\\
    \textbf{Output}: Best code $C^*$
    \begin{algorithmic}[1] 
        \STATE Initialize root $n_r$, $t{=}0$.
        \STATE Generate $N_I$ via $i$; link to $n_r$; evaluate all.
        \STATE $q_{\max}{=}-1e5$, $q_{\min}{=}0$.
        
        \WHILE{$t \le T$}
            \STATE $\lambda \leftarrow \lambda_0 * \frac{T-t}{T}$
            \STATE Updating $E$ if new top k heuristic functions emerge
            \STATE $n_c \leftarrow n_r$
            \WHILE{$n_c$ not leaf and $\text{depth}(n_c){<}H$}
                \STATE $n_c \leftarrow 
                \arg\max\limits_{c \in \text{Children}(n_c)}
                \Big(
                    \frac{Q(c)-q_{\min}}{q_{\max}-q_{\min}}
                    + \lambda \sqrt{{\ln(N(n_c)+1)}/{N(c)}}
                \Big)$
                \IF{ProgressiveWidening($n_c$)}
                    \STATE Expand $n_c$; simulate; backprop.
                    \STATE $t \leftarrow t+1$.
                \ENDIF
            \ENDWHILE
        
            \STATE $S_{\text{elite}} \leftarrow \text{Sample}(E,\text{real\_m})$
            \STATE $CCS \leftarrow n_c \cup S_{\text{elite}}$
            \STATE $E_k \leftarrow$ rapid cognition$(CCS)$
            \STATE $E_f \leftarrow$ complex cognition$(E_k,K^+,K^-)$
            \STATE Apply $\{em1,em2,m1,m2\}$ to $n_c$
            \IF{$C_{t}$}
                \IF{$S^{em1}_k=S^{em2}_m$} \STATE $K^- \leftarrow E_k$ 
                \ELSE
                \STATE $K^+ \leftarrow E_k$
                \ENDIF
            \ENDIF
            \STATE Evaluate new heuristics; update $Q,N$ and $C^*$
            \STATE $t \leftarrow t+8$
            \FOR{$c\in \text{Children}(n_c)$}
                \STATE $q_{\max}\leftarrow\max(q_{\max},Q(c))$, 
                $q_{\min}\leftarrow\min(q_{\min},Q(c))$
            \ENDFOR
        
            \WHILE{$n_c \ne n_r$}
                \STATE $Q(n_c)\leftarrow\max Q(c)$; $N(n_c)\leftarrow\ N(n_c)+8$
                \STATE $n_c \leftarrow \text{Father}(n_c)$
            \ENDWHILE
            \STATE $Q(n_r)\leftarrow\max Q(c)$; $N(n_c)\leftarrow\ N(n_c)+8$
        
        \ENDWHILE
        
        \STATE \textbf{Return} $C^*$.
    \end{algorithmic}
\end{algorithm}
\section{Methodology}
This section presents CogMCTS framework. Its overall structure is shown in Fig.~\ref{fig:vis1}. The framework fully preserves all the heuristics generated by the LLM in the MCTS decision tree. This enables systematic management and optimization of the generation process. The root node $n_{r}$ is virtual and does not correspond to any specific heuristic. All other nodes represent executable heuristics implemented in Python, with detailed descriptive annotations. The framework deeply integrates multi-round cognitive guidance from the LLM into the MCTS search. This enables the effective discovery of high-performance heuristics. During node expansion, the framework uses a dual-track strategy to balance exploration and exploitation. This approach demonstrates significant performance advantages in complex combinatorial optimization problems. It offers a robust and general solution for the efficient design of automated heuristic algorithms.

\subsection{Overall Architecture}
During initialization, CogMCTS generates $N_{I}$ initial nodes by providing a seed heuristic function. Multiple role settings are included in the prompts, such as an expert in heuristic optimization or virtual characters from different academic backgrounds. Each corresponding to a heuristic by the LLM. These nodes are arranged under a structural root node $n_{r}$. The root node serves only as an anchor for the search tree and has no heuristic meaning. The algorithm then repeats selection, expansion, simulation, and backpropagation. This continues until the number of heuristic evaluations reaches the limit $\mathrm{T}$.
\noindent\textbf{Selection}: In MCTS, the selection phase starts from the root node. The algorithm traverses down the tree using a tree policy to identify the most promising node for expansion. This process balances exploitation and exploration. In the selection phase of this framework, the UCT design adopts the exploration decay mechanism from MCTS-AHD. The formula is as follows:
\begin{equation}
\begin{gathered}
a^{*} = \arg\max_{a \in A(s)} \left[
\frac{Q(s,a) - q_{\min}}{q_{\max} - q_{\min}}
+ \lambda \sqrt{\tfrac{\ln (N(s)+1)}{N(c(s,a))}}
\right], \\[10pt]
\lambda = \lambda_{0} \cdot \frac{T - t}{T}.
\end{gathered}
\end{equation}
Here, $A(s)$ denotes the set of executable actions of node $s$. $q_{\mathrm{max}}$ and $q_{\mathrm{min}}$ are the maximum and minimum heuristic quality values $Q(\cdot)$ observed during the MCTS process. They are used to normalize the UCT calculation of child nodes. $N(s)$ denotes the number of visits to node $s$. $c(s,a)$denotes the child node generated by applying action $a$ at node $s$. For node selection, we adopt a progressive widening strategy, expanding high-visit nodes. The expansion actions and parameter settings follow those of MCTS-AHD for simplicity. An exploration decay mechanism is used, where $\lambda_{0}$=0.1.

\noindent\textbf{Expansion}: During the expansion phase, we propose a dual-track mechanism that combines MCTS with evolutionary ideas. As shown in Fig.~\ref{fig:vis2}, the i, em1, em2, m1, and m2 operations are designed. At nodes selected by UCT, the i operation provides seed functions to the LLM. Multiple-role prompts are used to generate a diverse set of initial heuristics as candidates for the subsequent search. The mechanism then expands along two complementary paths.
One path uses em1 and em2 to improve and iterate candidate heuristics based on historical feedback and global elite information. The other path uses m1 and m2 to introduce innovations in algorithm structure and parameter space, exploring new heuristic forms and increasing diversity.
The following sections detail the implementation of the five expansion operations.

\noindent\textit{i}: This operation initializes the heuristic population by providing task specifications to the generative LLM. To take advantage of existing knowledge and enhance diversity, the model receives seed heuristic functions and multiple-role prompts. Based on these inputs, the LLM generates $N_{I}$ heuristic functions with their functional descriptions. These serve as a high-quality and diverse candidate set for the subsequent expansion and optimization phases.

\noindent\textit{em1}: This operation performs local contrast–driven heuristic generation. Heuristic algorithms from the selected cognitive candidate set are paired. They are put into the LLM in batches, together with their performance metrics and textual descriptions. The systematic analysis and improvement cues from the cognitive guidance mechanism are appended to the prompts. The LLM uses contrast information and explicit improvement cues to identify and retain strong components. Then it addresses structural and methodological weaknesses to generate $m$ new heuristic functions.

\noindent\textit{em2}: This action implements global-best–guided heuristic generation. Specifically, the current global-best heuristic, its performance metrics, textual description, and global knowledge from the cognitive guidance mechanism are incorporated into the prompt. Based on this information, the LLM generates new heuristics to further enhance performance in the global search space.

\noindent\textit{m1}: This action performs a local structural mutation. In this step, the algorithm at the UCT-selected node is given to the LLM to generate a new algorithm. The new algorithm differs in form while improving the original. The operation aims to introduce novel mechanisms, formulas, or program modules to expand the design space and enhance potential performance.

\noindent\textit{m2}: This action performs parameter optimization and strategy fine-tuning. The algorithm at the UCT-selected node is given to the LLM. It guides the selection of key parameters and supports the creation of a new algorithm. The new algorithm differs in parameter settings and their interaction with the original formulas. This enables improvements and broader exploration.

\noindent\textbf{Simulation}: Then, CogMCTS evaluates the new heuristic ${h}$ on the dataset ${D}$ and calculates its reward ${R(h)}$. This reward is used directly as the quality value in UCT calculations $Q(s,a)\leftarrow R(h)$. Simultaneously, the reflective population, the global-best heuristic, the global-best value, and the global-worst value are updated in real time.

\noindent\textbf{Backpropagation}: During backpropagation, simulation results are passed from the leaf node ${s}$ to the root $n_{r}$ along the search path. The quality estimates $Q(\cdot)$ and visit counts $N(\cdot)$ of all nodes on this path are updated at each level. CogMCTS updates the quality values and visit counts as follows:
\begin{equation}
\begin{gathered}
Q(s,a) \leftarrow \max_{a \in A(s)} Q(s,a), \\[5pt]
N(s) \leftarrow \sum_{a \in A(s)} N(c(s,a))
\end{gathered}
\end{equation}

\begin{table*}[t]
\centering
\renewcommand\arraystretch{1.1}
\setlength{\tabcolsep}{1.2mm}
\resizebox{\textwidth}{!}{
\begin{tabular}{l|cccc|cccc|cccc}
\bottomrule[0.5mm]
            & \multicolumn{4}{c|}{OP} 
            & \multicolumn{4}{c|}{CVRP} 
            & \multicolumn{4}{c}{MKP} \\ \hline
Test sets   & \multicolumn{2}{c}{{$N$=50}} 
            & \multicolumn{2}{c|}{$N$=100} 
            & \multicolumn{2}{c}{{$N$=50, $C$=50}} 
            & \multicolumn{2}{c|}{$N$=100, $C$=50} 
            & \multicolumn{2}{c}{{$N$=100, $m$=5}} 
            & \multicolumn{2}{c}{$N$=200, $m$=5}\\
\midrule[0.3mm]
Methods     & Obj.↑ & Gap↓ & Obj.↑ & Gap↓
            & Obj.↓ & Gap↓ & Obj.↓ & Gap↓
            & Obj.↑ & Gap↓ & Obj.↑ & Gap↓ \\ \hline
ACO         & 13.354 & 32.78\% & 24.131 & 33.69\%
            & 11.355 & 27.77\% & 18.778 & 25.76\% 
            & 22.738 & 2.39\% & 40.672 & 4.40\%  \\
DeepACO     & \textbf{19.867} & \textbf{0.00}\% & \textbf{36.392} & \textbf{0.00}\%
            & \textbf{8.888} & \textbf{0.00\%} & \textbf{14.932} & \textbf{0.00\%} 
            & 23.093 & 0.86\% & 41.988 & 1.30\%  \\ \hline
\multicolumn{13}{c}{LLM-based AHD: \textit{GPT-4o-mini}}  \\ \hline
EoH         & 15.293& 23.02\% & 30.626  & 15.84\% 
            & 9.359 & 5.31\% & 15.681 & 5.02\% 
            & 23.139 & 0.67\% & 41.994 & 1.29\% \\
ReEvo       & 15.224 & 23.37\% & 30.303 & 16.73\%
            & 9.327 & 4.94\% & 16.092 & 7.77\% 
            & 23.245 & 0.21\% & 42.416 & 0.30\% \\
MCTS-AHD    & 15.186 & 23.56\% & 30.162 & 17.12\%
            & 9.286 & 4.48\% & 15.782 & 5.70\% 
            & 23.269 & 0.11\% & 42.498 & 0.11\% \\
CogMCTS(Ours) 
            & \cellcolor[HTML]{D9D9D9}{15.370} & \cellcolor[HTML]{D9D9D9}{22.63\%} 
            & \cellcolor[HTML]{D9D9D9}{30.814} & \cellcolor[HTML]{D9D9D9}{15.33\%} 
            & \cellcolor[HTML]{D9D9D9}9.239 & \cellcolor[HTML]{D9D9D9}3.95\% 
            & \cellcolor[HTML]{D9D9D9}15.594 & \cellcolor[HTML]{D9D9D9}4.43\% 
            & \cellcolor[HTML]{D9D9D9}\textbf{23.294} & \cellcolor[HTML]{D9D9D9}\textbf{0.00\%} 
            & \cellcolor[HTML]{D9D9D9}\textbf{42.542} & \cellcolor[HTML]{D9D9D9}\textbf{0.00\%} \\ 
\toprule[0.5mm]
\end{tabular}
}
\label{acoall}
\caption{Heuristics are designed within the general ACO framework to solve OP, CVRP, and MKP problems. Each test set contains 64 independent instances. Algorithm performance is evaluated by averaging results over three independent runs. The best-performing method under each LLM is highlighted in gray, and the overall best result for each test set is shown in bold.}
\end{table*}

\subsection{Cognitive-Guided Mechanism}
The CogMCTS framework uses a cognitive guidance mechanism. It enables systematic knowledge extraction and optimization during the search process. The mechanism has two complementary stages: rapid cognition and complex cognition. These stages provide multi-level feedback to guide heuristic evolution.

In rapid cognition, node functions are ranked by performance. They are then sequentially fed into an LLM as algorithm descriptions, performance metrics, and code. The LLM performs pairwise comparisons to generate analytical conclusions and extracts latent design patterns as empirical knowledge. In the complex cognition stage, the mechanism integrates knowledge from multiple sources. Specifically, it combines feedback from the rapid cognition stage, positive knowledge accumulated from previous iterations, and induced negative knowledge. This guides the LLM to generate structured guidance containing keywords, recommendations, avoidance cues, and explanations, enhancing the completeness and precision of heuristic design. The distinction between positive and negative knowledge relies on a Consistency-based Knowledge Validation (CKV) mechanism. CKV compares two global-best solutions. One $S_{em1}^k$ is obtained before executing em1 in the $k^\text{th}$ iteration. The other $S_{em2}^m$ is obtained after executing em2 in the $m^\text{th}$ iteration. If the solutions differ, empirical knowledge $E_{k}$ from the rapid cognition stage of the $k^\text{th}$ iteration is stored as positive knowledge $K^{+}$. If they are identical, they are stored as negative knowledge $K^{-}$. This prevents ineffective experience from accumulating in later searches. CKV can be formalized as:
\begin{equation}
 E_{k}\;\; \mapsto \;\;
\begin{cases}
\mathcal{K}^{+}, & \text{if } \Delta(S) = S_{em2}^{m} \ominus S_{em1}^{k} \neq \varnothing, \\
\mathcal{K}^{-}, & \text{if } \Delta(S) = S_{em2}^{m} \ominus S_{em1}^{k} = \varnothing.
\end{cases}
\end{equation}
Here, $\ominus$ denotes the operator representing the difference between global optimal solutions, $\Delta(S)$ indicating the set of variables in the solution space.

\begin{table}[t]
\centering
\renewcommand\arraystretch{1.05}
\setlength{\tabcolsep}{1.1mm}
\resizebox{0.48\textwidth}{!}{
\begin{tabular}{l|c|cccc} 
\bottomrule[0.5mm]
\multicolumn{6}{c}{\textbf{TSP}} \\ 
\midrule[0.3mm]
Methods & LLMs & {20} & {50} & {100} & {200} \\ 
\midrule[0.3mm]
KGLS   & \ding{55} & 0.004\% & 0.017\% & 0.002\% & 0.284\% \\
NeuOpt  & \ding{55} & 0.000\% & 0.000\% & 0.027\% & 0.403\% \\ 
GNNGLS  & \ding{55} & 0.000\% & 0.052\% & 0.705\% & 3.522\% \\ 
NeuralGLS  & \ding{55} & 0.000\% & 0.003\% & 0.470\% & 3.622\% \\ 
\hline
\multicolumn{6}{c}{LLM-based AHD: \textit{GPT-4o-mini}} \\ 
\hline
EoH        & \ding{51} & 0.015\% & 0.004\% & 0.004\% & 0.263\% \\
ReEvo      & \ding{51} & \cellcolor[HTML]{D9D9D9}\textbf{0.000}\% & {0.001}\% & 0.009\% & 0.278\% \\
MCTS-AHD     & \ding{51}& \cellcolor[HTML]{D9D9D9}\textbf{0.000}\% & 0.011\% & 0.005\% & 0.295\% \\
CogMCTS   & \ding{51}& {0.015\%} & \cellcolor[HTML]{D9D9D9}\textbf{0.000\%} & \cellcolor[HTML]{D9D9D9}\textbf{0.000\%} & \cellcolor[HTML]{D9D9D9}\textbf{0.242\%}  \\
\toprule[0.5mm]
\end{tabular}
}
\label{bpponline}
\caption{Heuristics are designed within the general GLS framework to solve TSP problems. The best-performing method under each LLM is highlighted in gray, and the overall best result for each test set is shown in bold.}
\end{table}

\subsection{Cognitive Candidate Set}
The Cognitive Candidate Set (CCS) serves as the core input for the cognitive guidance mechanism. It carries heuristic node information required for systematic analysis and extraction of experience. CCS consists of two types of nodes. The first type comprises high-potential nodes chosen from the current MCTS using the UCT strategy. These nodes indicate the main directions for exploration in the search tree. The second type contains up to ${m}$ nodes sampled from the global elite $E=\{x_1,x_2,\ldots,x_N\}$ set with non-uniform probabilities. The sampling strategy assigns weights according to node ranking $r_{i}$. After normalization, nodes are drawn without replacement, and up to ${m}$ nodes are added to the candidate set. High-quality nodes receive higher priority, while lower-ranked nodes are also included. This maintains diversity and novelty in the search space. By combining local high-potential nodes with global elite information, CCS provides an efficient multi-level input. It supports rapid cognition, complex cognition, and the evolution of systematic knowledge.
\begin{equation}
\begin{gathered}
\tilde{w}_i=\frac{1}{r_{i}+1+N}, \quad i=1,\ldots,N, \\[6pt]
CCS = S_{UCT}\cup \mathcal{S}_m^{ \begin{Bmatrix} p_i=\frac{\tilde{w}_i}{\sum_{j=1}^N\tilde{w}_j} \end{Bmatrix}_{i=1}^N}(E)
\end{gathered}
\end{equation}
Here, $\tilde{w}_i$ represents unnormalized weights. $S_{{UCT}}$ denotes high-potential nodes selected from the search tree using the UCT strategy. $\mathcal{S}_m^{\{p_i\}}(\cdot)$denotes up to ${m}$ elements sampled without replacement from the set ${E}$ according to a probability distribution $\{p_i\}_{i=1}^N$.

\begin{table*}[t] \centering \scriptsize \renewcommand\arraystretch{0.9} \setlength{\tabcolsep}{2.0mm} \resizebox{\textwidth}{!}{ \begin{tabular}{l|c|cc|cc|cc|cc} 
\bottomrule[0.5mm] 
\multirow{3}{*}{\raisebox{-3mm}{\raggedright Methods}} & 
\multirow{3}{*}{\raisebox{-3mm}{\centering LLMs}} &
\multicolumn{8}{c}{KP} \\ 
\cline{3-10} 
& & 
\multicolumn{2}{c|}{\raisebox{-1.5mm}{$N$=50, $W$=12.5}} & 
\multicolumn{2}{c|}{\raisebox{-1.5mm}{$N$=100, $W$=25}} & 
\multicolumn{2}{c|}{\raisebox{-1.5mm}{$N$=200, $W$=25}} & 
\multicolumn{2}{c}{\raisebox{-1.5mm}{$N$=500, $W$=25}} \\

\cline{3-10} 
& & \raisebox{-1.5mm}{Obj.↑} & \raisebox{-1.5mm}{Gap↓} 
& \raisebox{-1.5mm}{Obj.↑} & \raisebox{-1.5mm}{Gap↓} 
& \raisebox{-1.5mm}{Obj.↑} & \raisebox{-1.5mm}{Gap↓} 
& \raisebox{-1.5mm}{Obj.↑} & \raisebox{-1.5mm}{Gap↓} \\ 
\midrule[0.3mm]
 
OR-Tool & \ding{55} & \textbf{20.037} & \textbf{0.00}\% & \textbf{40.271} & \textbf{0.00}\% & \textbf{57.448} & \textbf{0.00}\% & \textbf{90.969} & \textbf{0.00}\% \\ 

Greedy Construct & \ding{55} & 19.985 & 0.260\% & 40.225 & 0.114\% & 57.395 & 0.092\% & 90.919 & 0.055\% \\ 

POMO & \ding{55} & 19.612 & 2.12\% & 39.676 & 1.48\% & 57.271 & 0.308\% & 85.580 & 5.92\% \\ \hline

\multicolumn{10}{c}{\raisebox{-0.3mm}{\hspace{15.5em}LLM-based AHD: \textit{GPT-3.5-turbo}}} \\ \hline

Funsearch & \ding{51} & 19.985 & 0.260\% & 40.225 & 0.114\% & 57.395 & 0.092\% & 90.919 & 0.055\%\\

EoH & \ding{51} & 19.994 & 0.215\% & 40.231 & 0.099\% & 57.400 & 0.084\% & 90.919 & 0.055\%\\ 

MCTS-AHD & \ding{51} & 19.996 & 0.205\% & \cellcolor[HTML]{D9D9D9}40.232 & \cellcolor[HTML]{D9D9D9}0.097\% & 57.393 & 0.096\% & 90.923 & 0.051\% \\ 

CogMCTS(ous) & \ding{51} & \cellcolor[HTML]{D9D9D9}19.997 & \cellcolor[HTML]{D9D9D9}0.200\% & \cellcolor[HTML]{D9D9D9}40.232 & \cellcolor[HTML]{D9D9D9}0.097\% & \cellcolor[HTML]{D9D9D9}57.403 & \cellcolor[HTML]{D9D9D9}0.078\% & \cellcolor[HTML]{D9D9D9}90.924 & \cellcolor[HTML]{D9D9D9}0.049\%   \\ \hline

\multicolumn{10}{c}{\raisebox{-0.3mm}{\hspace{15em}LLM-based AHD: \textit{GPT-4o-mini}}} \\ \hline

Funsearch & \ding{51} & 19.987 & 0.250\% & 40.225 & 0.114\% & 57.398 & 0.087\% & 90.920 & 0.054\%\\ 

EoH & \ding{51} & 19.993 & 0.220\% & 40.231 & 0.099\% & 57.392 & 0.097\% & 90.867 & 0.112\% \\

MCTS-AHD & \ding{51} & 19.995 & 0.210\% & 40.231 & 0.099\% & 57.399 & 0.085\% & 90.921 & 0.053\% \\ 

CogMCTS(ous) & \ding{51} & \cellcolor[HTML]{D9D9D9}19.996 & \cellcolor[HTML]{D9D9D9}0.205\% & \cellcolor[HTML]{D9D9D9}40.235 & \cellcolor[HTML]{D9D9D9}0.089\% & \cellcolor[HTML]{D9D9D9}57.405 & \cellcolor[HTML]{D9D9D9}0.075\% & \cellcolor[HTML]{D9D9D9}90.923 & \cellcolor[HTML]{D9D9D9}0.051\%\\ \hline
\toprule[0.5mm] 
\end{tabular} } 
\label{kp}
 \caption{Heuristics are designed under the step-by-step framework to solve the KP problem. Optimal for KP is the result of OR-Tools. Each test set contains 1,000 instances. Algorithm performance is evaluated by averaging results over three independent runs. The best-performing method under each LLM is highlighted in gray, and the overall best result for each test set is shown in bold.}
\end{table*}

\section{Experiments}
This section evaluates the performance of the proposed CogMCTS method on several NP-hard combinatorial optimization problems. The test problems include the Orienteering Problem (OP), the Capacitated Vehicle Routing Problem (CVRP), and the Multiple Knapsack Problem (MKP) under the ACO framework, Traveling Salesman Problem (TSP) task under the GLS framework, as well as the KP under a step-by-step construction framework.

\noindent\textbf{Settings:} For experiments in this section, we set the initial number of tree nodes $N_{I}$ to 10 and the parameter $\lambda_{0}$ to 0.1. We limited each heuristic function to 60 seconds of runtime on its respective evaluation dataset ${D}$. To ensure the LLM-based AHD method works effectively and flexibly across different pretrained models, we validated it using both GPT-3.5-turbo and GPT-4o-mini.

\noindent\textbf{Baselines:} To systematically evaluate the effectiveness of CogMCTS in heuristic function design, this study selects three representative types of baseline methods for comparison: (1) Manually designed heuristics, such as ACO for MKP, CVRP, and OP ~\cite{dorigo2007ant}, Greedy-Construct (GC)  for KP; (2) NCO methods, such as POMO ~\cite{kwon2020pomo} and DeepACO ~\cite{ye2023deepaco}; (3) LLM-based AHD methods, including FunSearch, EOH, ReEvo, and the recently proposed MCTS-AHD. To ensure fair comparison, we applied unified settings for CogMCTS and all LLM-based AHD baselines. We gave CogMCTS, FunSearch, ReEvo the same initial seed functions in each design scenario. We tested their performance using the same training datasets and evaluation procedures. We set the evaluation budget for all LLM-based AHD methods in the dataset ${D}$ to $T = 1000$. To further reduce statistical bias, we ran each LLM-based AHD method three times independently for each application scenario.

\subsection{Overall Results}

\noindent\textbf{Ant Colony Optimization Framework}: ACO is a probabilistic search method inspired by ant foraging. It combines heuristic information with pheromones to define solution construction probabilities. Alternating between solution generation and pheromone update, ACO balances exploration and exploitation, gradually guiding the search toward high-quality solutions ~\cite{dorigo2007ant,kim2024ant}. LLM-based AHD generates heuristic matrix functions to provide ACO with problem-specific construction preferences. This extends ACO into a flexible framework for diverse combinatorial optimization tasks and enables automated heuristic design and optimization. Using this framework, CogMCTS conducts heuristic design experiments on several NP-hard problems. The experiments include the OP, the  MKP, and the CVRP.

\noindent\textit{OP}: We evaluated the heuristic design capability of CogMCTS in OP within the ACO framework. CogMCTS first learned on a training set of 5 instances, each with $N = 50$ nodes. We then tested it on two sets of different sizes:$N = 50$ and $N = 100$ nodes, with 64 independent instances per size. We fixed the number of ants and iterations at $N_{ANTS}=20$ and $N_\text{ITERATIONS}=50$ for all experiments to ensure comparability and stability. The results show that CogMCTS consistently outperformed almost all test instances. Its performance exceeded both manually designed heuristics and existing LLM-based AHD methods, including EOH, ReEvo, and MCTS-AHD.

\noindent\textit{MKP}: We evaluated the heuristic design capability of CogMCTS in the MKP within the ACO framework. CogMCTS first trained on a set of 5 instances, each with $N = 100$ items. We set the number of ants to $N_{ANTS}=10$ and the number of iterations to $N_\text{ITERATIONS}=50$. We then tested its generalization on two sets of instances with $N = 100$ and $N = 200$ items, each containing 64 independent instances. During testing, we fixed the number of ants at $N_{ANTS}=10$ and increased iterations to $N_\text{ITERATIONS}=100$ to better assess performance on larger problems. The results show that CogMCTS consistently outperformed almost all test instances. It exceeded both manually designed heuristics and existing LLM-based AHD methods, including EOH, ReEvo, and MCTS-AHD.

\noindent\textit{CVRP}: We evaluated the heuristic design capability of CogMCTS in CVRP within the ACO framework. The algorithm was first trained on a set of 10 instances, each with $N = 50$ nodes and capacity $N_\text{capacity}=50$. During training, we set the number of ants to $N_{ANTS}=30$ and the number of iterations to $N_\text{ITERATIONS}=100$ to ensure comparability and stability. Next, we tested the algorithm on three instance sets with $N = 50$ and $N = 100$, each containing 64 independent instances and capacity $N_\text{capacity}=50$. We set the number of ants to $N_{ANTS}=30$ and iterations to $N_\text{ITERATIONS}=500$ for testing. Results show that the algorithm performed well on all test sets. It handled different CVRP scales effectively and provided a reliable basis for further heuristic optimization and strategy design.

\begin{table}[t]
\centering
\renewcommand\arraystretch{1.4}

\resizebox{\columnwidth}{!}{%
\begin{tabular}{l|cc|ccc}
\bottomrule[0.5mm]
\multirow{3}{*}{\raisebox{-3mm}{\raggedright $C_\text{t}$}} 
            & \multicolumn{2}{c|}{OP} 
            & \multicolumn{3}{c}{KP} \\ \cline{2-6}
            & $N=50$ & $N=100$ 
            & $N=50,\,W=12.5$ & $N=100,\,W=25$ & $N=200,\,W=25$ \\ \cline{2-6}
            & \multicolumn{2}{c|}{Obj.↑} 
            & \multicolumn{3}{c}{Obj.↑} \\ \midrule[0.3mm]
1 
            & {15.328} & {30.596} 
            & {19.996} & {40.232} & {57.401} \\ 
2 
            & \textbf{15.370} & \textbf{30.814} 
            & \textbf{19.996} & \textbf{40.235} & \textbf{57.405} \\
3 
            & {15.230} & {30.385} 
            & {19.987} & {40.225} & {57.399} \\
\toprule[0.5mm]
\end{tabular}%
}
\caption{To determine the value of $m-k$, tests are conducted on both OP and KP tasks. The large language model used is GPT-4o-mini. Each LLM-based AHD method is run three times, and the average performance is reported. The overall best result for each test set is highlighted in bold.}
\label{aco_op_kp}
\end{table}

\begin{table}[t]
\centering
\renewcommand\arraystretch{1.2}
\resizebox{\columnwidth}{!}{%
\begin{tabular}{l|cc|cc}
\bottomrule[0.5mm]
\multirow{3}{*}{\raisebox{-3mm}{\raggedright $C_\text{t}$}} 
            & \multicolumn{2}{c|}{MKP} 
            & \multicolumn{2}{c}{CVRP} \\ \cline{2-5}
            & $N=100,\,m=5$ & $N=200,\,m=5$ 
            & $N=50,\,C=50$ & $N=100,\,C=50$ \\ \cline{2-5}
            & \multicolumn{2}{c|}{Obj.↑} 
            & \multicolumn{2}{c}{Obj.↓} \\ \midrule[0.3mm]
1 
            & {23.260} & {42.529} 
            & {9.310} & {15.782} \\ 
2 
            & \textbf{23.294} & \textbf{42.542} 
            & \textbf{9.239} & \textbf{15.594} \\ 
3 
            & {23.286} & {42.516} 
            & {9.242} & {15.639} \\ 
\toprule[0.5mm]
\end{tabular}%
}
\caption{To determine the value of $m-k$, tests are conducted on both MKP and CVRP tasks. The large language model used is GPT-4o-mini. Each LLM-based AHD method is run three times, and the average performance is reported. The overall best result for each test set is highlighted in bold.}
\label{aco_mkp_cvrp}
\end{table}

\noindent\textbf{Guided Local Search}:
The GLS framework is an iterative solution paradigm widely used in combinatorial optimization (CO). Its core mechanism iterates solutions through local search algorithms, such as the 2-opt operator. A penalty mechanism is introduced to guide the search process and help escape local optima. This method relies on manually designed key heuristic functions. These functions ensure solution feasibility and continuously improve solution quality during the search. As a general solving framework, GLS provides a theoretical foundation for LLM-based AHD methods. It is also a commonly used strategy for various COPs.

\noindent\textit{TSP}: In the context of the TSP, we systematically evaluated the heuristic design capability of CogMCTS using the GLS framework. CogMCTS was first trained on a dataset containing 10 instances, each with
$N = 200$. Its generalizability was then tested in four sets of instances with different scales:$N = 20$, $N = 50$, $N = 100$, $N = 200$, each set containing 64 independent instances. This setup allows for a comprehensive assessment of CogMCTS in varying problem sizes. The optimal solutions were obtained using the LKH algorithm. The results show that CogMCTS consistently outperforms other methods in multiple test instances.

\noindent\textbf{Step-by-Step Construction Framework}: The step-by-step  construction framework is an iterative algorithm paradigm widely used in combinatorial optimization. Its core idea is to build a complete solution through multiple sequential decisions. Each step relies on heuristic information or guidance from a learning model. This approach ensures the feasibility of the solution while continuously improving the quality of the solution. As a general solving paradigm, it forms the foundation for LLM-based AHD methods. It is also one of the most common solutions in NCO ~\cite{vinyals2015pointer,bello2016neural}.

\noindent\textit{KP}: In the context of the KP, we systematically evaluated CogMCTS’s heuristic design ability using the Step-by-Step Construction framework. CogMCTS first trained on a set of 64 instances, each with $N = 100$ items represented by 100 two-dimensional coordinates in the [0,1] range. We then tested its generalization on four sets of instances with $N = 50$, $N = 100$, $N = 200$ and $N = 500$, each containing 1000 independent instances. This setup allowed a comprehensive assessment of CogMCTS across different problem scales. The baseline GC method selects items in KP based on the value-to-weight ratio. The results show that CogMCTS consistently outperforms almost all test instances. It exceeds both manually designed greedy heuristics and existing LLM-based AHD methods, including Funsearch, EoH, and MCTS-AHD. In the KP task experiment using the ReEvo framework, the same seed heuristic was applied for initialization and evolution. However, the system produced only one heuristic individual. The evolution stopped early and no data was obtained.

\begin{table}[t]
\centering
\renewcommand\arraystretch{1.2}
\resizebox{\columnwidth}{!}{%
\begin{tabular}{c|cccc|c} 
\bottomrule[0.5mm]
\multirow{2}{*}{Task} 
            & \multicolumn{4}{c|}{Choice} 
            & Gap \\ \cline{2-5}
            & action em1 & action em2 & action m1 & action m2 & \multicolumn{1}{c}{} \\ \midrule[0.3mm]
\multirow{4}{*}{\centering KP100} 
            &  &  & \checkmark & \checkmark & 0.117\%   \\ 
            & \checkmark &  & \checkmark & \checkmark & 0.097 \%\\ 
            &  & \checkmark & \checkmark & \checkmark & 0.102 \%\\ 
            & \checkmark & \checkmark & \checkmark & \checkmark & \textbf{0.089}\%\\ 
            
\toprule[0.5mm]
\end{tabular}%
}
\caption{We conducted an ablation analysis on the actions of CogMCTS. Different variants were used to generate heuristics within the step-by-step construction framework. The optimality gaps were evaluated on 1,000 test instances. All results were averaged over three runs to ensure reliability.}
\label{tab:kp100_actions}
\end{table}

\begin{figure}[t]
  \centering
  \adjustbox{trim=0.8cm 0.2cm 1.0cm 0.4cm, clip}{
    \includegraphics[width=\columnwidth]{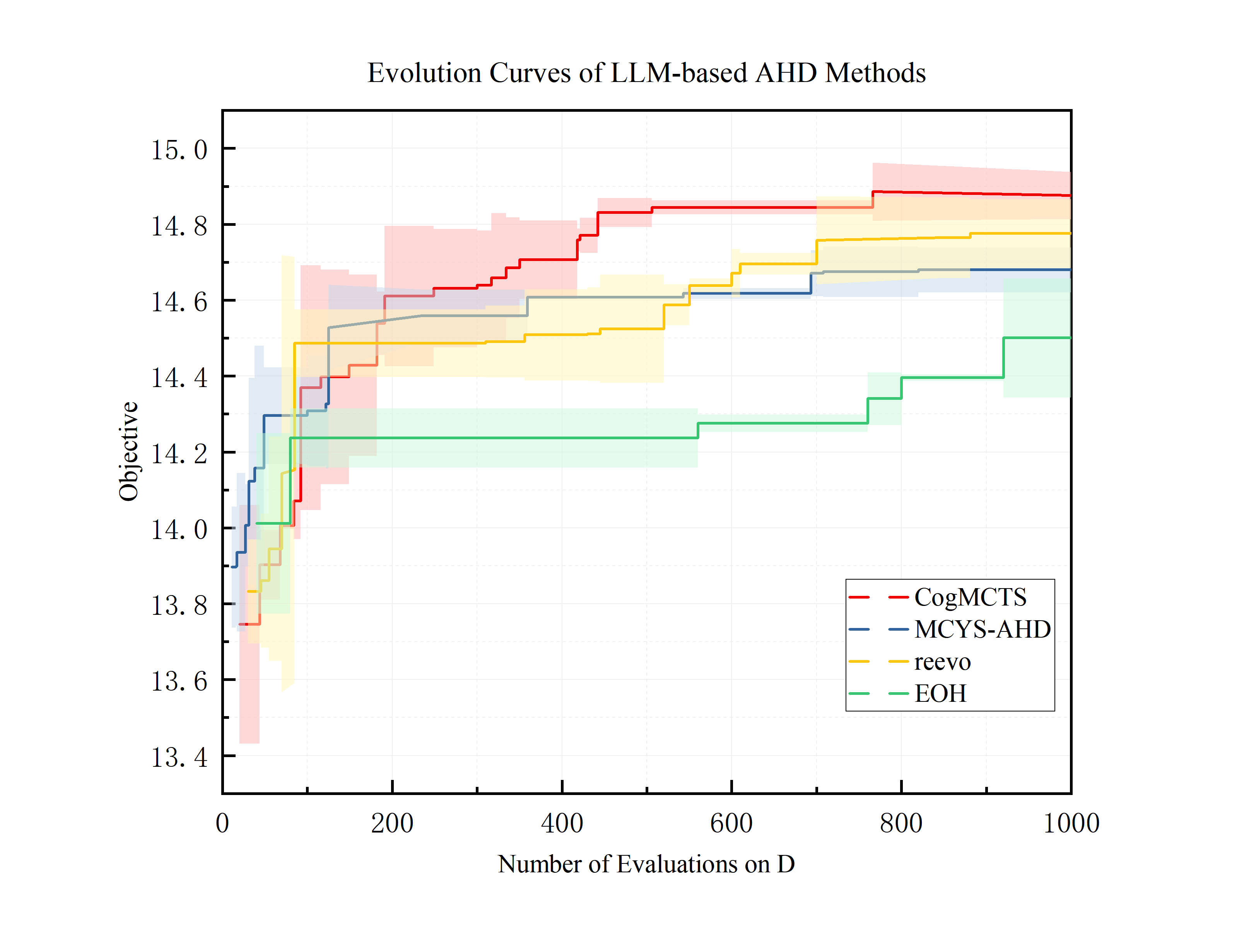}
  }
  \setlength{\abovecaptionskip}{-0.15cm} 
  \caption{Evolution curves on the OP illustrating convergence performance}
  \label{fig:vis3}
\end{figure}

\section{Ablation Study}
In the CKV mechanism, the interval $m-k$ between the $k^\text{th}$ em1 and the $m^\text{th}$
em2 defines a thinking cycle ($C_\text{t}$). At the end of each cycle, we compare the global-best solution $S_{em1}^{k}$ before the $k^\text{th}$ em1 execution with the global-best solution $S_{em2}^{m}$ after the $m^\text{th}$ to assess the value of the experience from the $k^\text{th}$ fast reflection. If the global optimum changes, the experience is stored in the positive knowledge $K^{+}$ base; otherwise, it is stored in the negative knowledge $K^{-}$ base. Experiments on OP, MKP, CVRP, and KP show that the mechanism performs best when the $C_\text{t}=2$ (see Tables~\ref{aco_op_kp} and \ref{aco_mkp_cvrp}). This parameter is critical. A two-round window length balances exploration and verification. It avoids early judgment that could discard useful experience and prevents noise from excessively long intervals. This setting significantly improves the discovery rate of global-best solutions and enhances the quality of the knowledge base. Thus, it represents the optimal configuration for the consistency-driven knowledge verification mechanism.

To validate the effectiveness of the cognitive guidance mechanism combined with MCTS, we conducted systematic ablation experiments on node expansion actions. According to Table~\ref{tab:kp100_actions}, in the KP task, removing em1, em2, or both em1 and em2 led to MCTS variants that could only generate low-performing heuristics. The impact was most significant when both em1 and em2 were removed. These results clearly demonstrate that the cognitive guidance mechanism plays a key role in heuristic generation and optimization in CogMCTS. It also significantly improves the performance of the algorithm.

We visualized the training curves for the OP task in Fig.~\ref{fig:vis3}. The results show that the CogMCTS framework can generate high-performance solutions quickly, stably, and continuously in the heuristic function design task.

\section{Conclusion}
This paper proposes the CogMCTS framework, which integrates a cognitive guidance mechanism into MCTS. By leveraging multi-round feedback from LLMs, the framework strengthens high-quality heuristics and gradually refines less effective heuristics. Structured feedback preserves the relevance and quality of the generated heuristics. Strategic node expansion combined with elite heuristic management balances exploration and exploitation. Experimental results show that CogMCTS outperforms existing LLM-based adaptive heuristic methods in stability, efficiency, and discovery of high-quality solutions. It demonstrates clear advantages in solving complex combinatorial optimization problems. For future work, it is promising to improve the efficiency of the cognitive guidance mechanism and explore lightweight simulation strategies such as Gumbel-MCTS to accelerate large-scale optimization. And it is also encouraged to extend the framework to more complex combinatorial and real-world optimization problems.

\section*{Ethical Statement}

There are no ethical issues.

\section*{Acknowledgments}
This work was financially supported by the 2024 Major Science and Technology Project of Hefei (Project No. 2024SZD006), the National Natural Science Foundation of China (Key Project, Grant No. 62236002).

\bibliographystyle{named}
\bibliography{ijcai26}

\end{document}